\documentclass[acmsmall]{acmart}

\usepackage{enumitem}
\usepackage{libertine}
\usepackage{algorithm}
\usepackage{algpseudocode}
\usepackage{amsmath}

\usepackage{amssymb}
\usepackage{microtype}
\usepackage{graphicx}
\usepackage{booktabs}
\usepackage{hyperref}
\usepackage{rme-macros}  % Richard Everson's macros.
\usepackage[caption=false,font=footnotesize]{subfig}
\usepackage{multirow}
\usepackage{siunitx}
\usepackage{etoolbox}
\usepackage{pdflscape}
\usepackage{colortbl}
\usepackage{makecell}

\newcommand{\sweetspot}{\Delta}  % Unconstrained uncertainty set.
\newcommand{\SWEETSPOT}{{\sweetspot}}  % Parameterised uncertainty set.
\renewcommand{\S}{\SWEETSPOT}  % Shorthand.
\newcommand{\robustset}{\Delta_\epsilon}
\newcommand{\domain}{\mathcal{X}}
\newcommand{\nbhd}{\mathcal{N}}
\newcommand{\bestx}{\bx^\star}

\newcommand{\prob}{p}
\newcommand{\fhat}{\hat{f}}
\newcommand{\xnext}{\bx_{new}}

\renewcommand{\defn}{\triangleq}
\renewcommand{\d}{\text{d}}

\renewcommand{\mnote}[2][\textcolor{red}{\dagger}]{$#1$\marginpar{\color{red}\raggedright\tiny$#1$#2}}

\newcommand{\beginsupplement}{%
        \setcounter{table}{0}
        \renewcommand{\thetable}{S\arabic{table}}%
        \setcounter{figure}{0}
        \renewcommand{\thefigure}{S\arabic{figure}}%
        \setcounter{section}{0}
        \renewcommand*{\thesection}{S\arabic{section}}%
        \clearpage%
     }

\setcopyright{acmcopyright}
\copyrightyear{2021}
\acmYear{2021}
\acmDOI{00.0000/0000000.0000000}
\acmJournal{JACM}
\acmVolume{0}
\acmNumber{0}
\acmArticle{000}
\acmMonth{0}

\begin{document}

%%
%% Front matter
%%

\title{Bayesian Search for Robust Optima}

\author{Nicholas~D.~Sanders}
\email{n.sanders@exeter.ac.uk}
\orcid{0000-0002-3603-1824}
\author{Richard~M.~Everson}
\email{r.m.everson@exeter.ac.uk}
\orcid{}
\author{Jonathan~E.~Fieldsend}
\email{j.e.fieldsend@exeter.ac.uk}
\orcid{0000-0002-0683-2583}
\affiliation{%
  \institution{The University of Exeter}
  \streetaddress{North Park Road}
  \city{Exeter}
  \state{England, UK}
  \postcode{EX4 4QF}
}
\author{Alma~A.~M.~Rahat}
\email{a.a.m.rahat@swansea.ac.uk}
\orcid{0000-0002-5023-1371}
\affiliation{
  \institution{Swansea University}
  \streetaddress{Fabian Way, Skewen}
  \city{Swansea}
  \state{Wales, UK}
  \postcode{SA1 8EN}
}

\begin{abstract}
  Many expensive black-box optimisation problems are sensitive to their inputs.
  In these problems it makes more sense to locate a region of good designs,
  than a single---possibly fragile---optimal design.

  Expensive black-box functions can be optimised effectively with Bayesian
  optimisation, where a Gaussian process is a popular choice as a prior over
  the expensive function. We propose a method for robust optimisation using
  Bayesian optimisation to find a region of design space in which the expensive
  function's performance is relatively insensitive to the inputs whilst retaining a good
  quality. This is achieved by sampling realisations from a Gaussian process
  that is modelling the expensive function, and evaluating the improvement for
  each realisation. The expectation of these improvements can be optimised
  cheaply with an evolutionary algorithm to determine the next location at
  which to evaluate the expensive function. We describe an efficient process to
  locate the optimum expected improvement.  We show empirically that evaluating
  the expensive function at the location in the candidate uncertainty region
  about which the model is most uncertain, or at random, yield the best convergence
  in contrast to exploitative schemes.  % TODO(nick) Check this paragraph

  We illustrate our method on six test functions in two, five, and ten
  dimensions, and demonstrate that it is able to outperform two
  state-of-the-art approaches from the literature. We also demonstrate our
  method one two real-world problems in 4 and 8 dimensions, which involve
  training robot arms to push objects onto targets. 
\end{abstract}

% TODO(nick) Update CCSXML section with all relevant concepts. See Alma's paper

\maketitle

%%
%% Introduction.
%%

\section{Introduction}
\label{sec:introduction}

Optimisation is the search for the best-performing design with respect to a
predefined objective function. In an ideal scenario the objective function
would be well behaved and insensitive to small changes in the design
parameters. Any loss of performance due to the mis-specification of the
parameters or the inaccuracy of the model itself would be negligible and
largely go unnoticed.  But, for most real-world problems this is not the case.
Often the landscape of real objective functions varies rapidly as the design
parameters change, so even a small perturbation to these parameters could lead
to an unacceptably diminished performance. Such perturbations are frequently
the manifestation of uncertainties in the design process, and arise for a
number of reasons. For example: tolerances in the manufacturing process may
mean that the realised design is slightly different from the modelled one; once
the design has been produced it may be operated away from its design
conditions; the model used to optimise the design may not accurately reflect
reality; or there may be uncertainties in the environmental conditions (e.g.
air temperature or pressure).

The goal of \textit{robust optimisation} is to locate the best designs
that have stable performance irrespective of small perturbations to the design
parameters. Classic global optimisers are often ineffective at solving this
problem, because robust optima do not necessarily coincide with global optima;
in fact optimal robust designs potentially exist in a different region of the
domain altogether!

A straightforward way to quantify the \textit{robustness} of a proposed optimum
is to evaluate the objective function for a large number of design parameters
in the vicinity of the optimum. However, this strategy is ineffective when the
objective function is expensive to evaluate. These expensive-to-evaluate
functions are common to many disciplines, including tuning the parameters of
machine learning algorithms \cite{bergstra2011algorithms, snoek2012practical},
robotics \cite{tesch2011using, lizotte2007automatic}, and other engineering
design problems \cite{daniels2018suite, anthony2003robust, wiesmann1998robust,
volz2019single, shourangiz2020state}. \textit{Bayesian optimisation} is a
principled and efficient technique for the global optimisation of these sorts
of functions. The idea underlying Bayesian optimisation is to place a prior
distribution over the objective function and then update that prior with
observations of the objective function (obtained by expensively evaluating it)
in order to yield a posterior predictive distribution.  This posterior
distribution thus encodes the optimiser's knowledge of the objective function
landscape. It is then used to inform where to make the next observation of the
objective function through the use of an acquisition function, which balances
the exploitation of regions known to have good performance with the exploration
of regions where there is little information about the function's response. A
Gaussian process is a popular choice of prior, because they are intuitive to
understand, capable of modelling the objective function accurately with few
data, and cheap to evaluate.

% Contribution.
The chief contribution of this paper is the introduction of a novel acquisition
function for the Bayesian robust optimisation of expensive black-box functions.
Although we have phrased robust optimisation as seeking an optimum robust to
``small perturbations'' of design parameters, the technique we present is
applicable to arbitrarily large perturbations.  We evaluate the method on 6
benchmark functions and a real-world problem showing that it provides
state-of-the-art performance compared with two competing algorithms.

%
% I think we are going to remove the `higher dimensions` section, so I've
% commented this contribution. -- Nick.
%%
% Secondly, we describe an efficient algorithm by which to compute the
% acquisition function in higher dimensional spaces.
%%

% Road map.
We begin by outlining background material and reviewing similar techniques in
Section~\ref{sec:background}.
Section~\ref{sec:bayesian-search-for-a-robust-optimum} builds upon the previous
section by introducing the Bayesian optimisation of the robust domain, and
giving a demonstration on a toy function in one dimension.  Results of five-
and ten-dimensional test problems are presented alongside analysis in
Section~\ref{sec:results}, where we also evaluate the algorithms on two active
learning robot pushing problems. Finally, the conclusion and suggestions for
future work can be found in Section~\ref{sec:conclusion}.

%%
%% Background.
%%

\section{Background}
\label{sec:background}

This section comprises background material in Bayesian optimisation (Section~
\ref{subsec:bayesian-optimisation}), Gaussian processes (Section~
\ref{subsec:gaussian-processes}), and robust optimisation for
expensive-to-evaluate functions (Section~\ref{subsec:robust_optimisation}).

\subsection{Bayesian optimisation}
\label{subsec:bayesian-optimisation}

Although stochastic search algorithms, such as evolutionary algorithms, have
been popular for the optimisation of black-box functions, Bayesian optimisation
is often more attractive, particularly for expensive-to-evaluate functions.
Through explicitly modelling the expensive function and accounting for the
uncertainty in the model, the search can be guided efficiently to promising
areas of the decision space: either those with high certainty of being better
than the current best solution, or those with high uncertainty that may be
better than the current best. See \cite{brochu2010tutorial} for an introduction
to Bayesian optimisation, and \cite{shahriari2016taking} for a recent
comprehensive review.

To be definite and without loss of generality, we assume that the goal of
optimisation is to minimise a function $f(\bx)$, where $\bx$ are the design
parameters in the feasible space $\domain \subset \reals^D$.

Bayesian optimisation relies on constructing a probabilistic model of $f(\bx)$.
Assume that $f(\bx)$ has been (expensively) evaluated at $N$ locations
$\{\bx_n\}_{n=1}^N$ so that data $\data = \{(\bx_n, f_n \defn
f(\bx_n))\}_{n=1}^N$ are available from which to learn a model. Then Bayesian
modelling is used to construct a posterior predictive distribution $p(f \given
\bx, \data)$ at any desired location $\bx$. Crucially, Bayesian modelling gives
not only a prediction of the function value at $\bx$, but the posterior
distribution quantifies the uncertainty in the prediction as well. Where next
to expensively evaluate in  $\domain$ is determined by an \textit{acquisition
function}, which balances the exploitation of predicted good values of $f(\bx)$ with
exploring uncertain and potentially good regions. Here  we use the popular
\textit{expected improvement} \cite{jones1998efficient}, which has been shown
to be effective in practice and for which some theoretical guarantees exist
\cite{bull2011convergence}. Alternatives such as the probability of improvement
\cite{kushner1964new} or upper-confidence bound \cite{srinivas2010gaussian,
brochu2010tutorial} could also be used.

If $f(\bx)$ is modelled to take the value $\fhat(\bx)$, then the
\textit{improvement} at $\bx$ is defined as
\begin{equation}
  \label{eq:improvement}
  I(\bx) = \max\left(f^\star - \fhat(\bx), 0\right) \,,
\end{equation}
where
\begin{equation}
  \label{eq:fstar}
  f^\star = \min_{\bx_n \in \data} f(\bx_n) = \min_n f_n
\end{equation}
is the best function value from the evaluations thus far. The expected
improvement is then
\begin{equation}
  \label{eq:expected_improvement}
  EI(\bx; \data) =
  \int_{-\infty}^{\infty} I(\bx) \prob(\fhat\given \bx, \data)
  \, \text{d}\fhat(\bx)
  \,.
\end{equation}
Gaussian processes are commonly used for modelling $f(\bx)$ in which case the
posterior predictive distribution is itself a Gaussian density (see Section
\ref{subsec:gaussian-processes}) with mean $\mu(\bx)$ and variance
$\sigma^2(\bx)$. In this case the expected improvement has the closed
analytical form \cite{jones1998efficient}:
\begin{equation}
  \label{eq:ei_analytical}
  EI(\bx; \data) = (f^\star - \mu(\bx))\Phi(Z) + \sigma(\bx)\phi(Z)
  \,,
\end{equation}
where $Z = (f^\star - \mu(\bx))/\sigma(\bx)$ and $\phi(\cdot)$ and
$\Phi(\cdot)$ are the standard Normal density and cumulative distribution
functions respectively.

The next (expensive) evaluation is then chosen as that with the greatest
expected improvement: $\bx' = \argmax_{\bx \in \domain} EI(\bx)$.  This
location is often discovered by using an evolutionary algorithm to maximise
$EI(\bx)$, which is rapid since $EI(\bx)$ is computationally cheap to evaluate.
The evaluated location and its function value are added to $\data$ and the
optimisation proceeds iteratively until some stopping criterion is met or, more
commonly, the available computational resources are exhausted.

\subsection{Gaussian processes}
\label{subsec:gaussian-processes}

Gaussian processes (GPs) \cite{rasmussen2006gaussian} are commonly used for
Bayesian optimisation due to their flexibility and the simple Gaussian
predictive posterior distributions. Briefly, a GP is a collection of random
variables, and any finite number of these have a joint Gaussian distribution.
Given data $\data$ and a feature vector $\bx$, the GP posterior predictive
density of the target $\fhat$ is Gaussian:
\begin{align}
  \label{eq:predictive}
  p(\fhat \given \bx, \data) = \normal(\fhat \given \mu(\bx), \sigma^2(\bx))\,,
\end{align}
where the mean and variance of the prediction are given by
\begin{align}
  \mu(\bx)      &= \bff\trp  K^{-1} \bk, \label{eq:pred-mean} \\
  \sigma^2(\bx) &=  k(\bx, \bx) - \bk\trp \bK^{-1} \bk \label{eq:pred-variance}
  \,.
\end{align}
Here $\bff = (f_1, f_2, \ldots, f_N)\trp $ is the vector of evaluated function
values at $\bx_1, \bx_2, \ldots, \bx_N$.  Non-linearity in the GP enters
through a kernel function $k(\bx, \bx'),$ which models the covariance between
two feature vectors. The $N \times N$ covariance matrix $\bK$ collects these
covariances together, $K_{ij} = k(\bx_i, \bx_j)$, and $\bk = \bk(\bx)$ is the
$N$-dimensional vector of covariances between the training data and $\bx$: $k_n
= k(\bx, \bx_n)$.  There are a number of kernels that could be used, for
example the squared exponential function or the Mat\'ern family of covariance functions
\cite{rasmussen2006gaussian}; here we used the Mat\'ern covariance function
with smoothing parameter $\nu=5/2$ which has been  recommended for modelling
realistic functions \citep{snoek2012practical}.

In addition the kernel function depends upon a number of hyper-parameters,
$\eta$.  Training the GP comprises inferring these hyper-parameters by
maximising the marginal likelihood  of the data $p(\data \given \eta)$ given by
\begin{align}
  \label{eq:log-marginal-likelihood}
  \log p(\data \given \eta) =
  - \frac{1}{2} \log |\bK|
  - \frac{1}{2} \bff\trp \bK\inv \bff
  - \frac{N}{2}\log (2\pi) \,.
\end{align}
Although the log marginal likelihood function landscape may be non-convex and
multi-modal, we adopt the standard practice of using a gradient-based optimiser
(L-BFGS) with several random starts to estimate good hyper-parameter values
\cite{gpy2012gaussian}.

\subsection{Robust Optimisation of Expensive Functions}
\label{subsec:robust_optimisation}

% TODO (nick) Add references to Beland (2017), Frolich (2020), Oliveira (2019).

The focus of robust optimisation is to determine function optima in the face of
uncertainty. These uncertainties are typically considered to arise in  one of
three categories: \textit{(a)} design uncertainties, \textit{(b)} model
uncertainties, or \textit{(c)} environmental uncertainties
\cite{beyer2007robust}. A variety of methods exist to alleviate the added
complexity of searching for robust optima; comprehensive surveys are given
by \citet{beyer2007robust} and \citet{gabrel2014recent}.

In this work we focus on one category of robustness: uncertainties or
mis-specification in the design parameters. In this case robust optimisation
seeks to find an $\bx \in \domain$ where some small perturbation $\bdelta$ does
not cause the objective function's response at $\bx + \bdelta$ to become
unacceptably poor. We define the set of all possible perturbations  to be
\begin{align}
  \label{eq:robust_set_definition}
  \robustset \triangleq
  \{ \bdelta \in \domain \given d(\bdelta) \leq \epsilon \}
\end{align}
where $\epsilon$ is the \textit{robustness parameter}, and $d(\cdot)$ is a
function controlling the shape of the robust region, which is often a
distance function. The choice of $\epsilon$ and $d(\cdot)$ is determined by the
problem owner.

A quality function $Q(\bx, \robustset)$ can be used to quantify the robustness
of $\bx$. There are a number of ways one could formulate $Q$, for example the
average performance over the perturbations: 
\begin{equation}
  \label{eq:quality-average}
  Q_{avg}\left( \bx,\, \robustset \right) =
  \int_{\robustset} f(\bx + \bdelta)  \pi(\bx + \bdelta) \, \d{\bdelta},
\end{equation}
where $\pi(\bx + \bdelta)$ denotes the probability that in practice $f(\cdot)$
will be evaluated at $\bx + \bdelta$ rather than $\bx$. Although for analytic
convenience $\robustset$ is sometimes taken to be unbounded and $\pi(\bx)$ is
taken to be a Gaussian distribution centred on $\bx$, in most real applications
$\robustset$ is taken to be a (small) compact set and $\pi$ is uniform on it.

An alternative quality function, which may be more useful in practice, is to
guarantee the worst-case performance \cite{beyer2007robust},
\begin{equation}
  \label{eq:quality-worst}
  Q_{max}\left(\bx,\, \robustset\right) =
  \sup_{\bdelta \in \robustset} f(\bx + \bdelta) \,.
\end{equation}
Having defined a quality function, robustness parameter, and shape function the
robust optimisation problem may be written as:
\begin{equation}
    \label{eq:unconstrained-optimisation}
    \min_{\bx \subset \domain} Q\left(\bx,\, \robustset\right).
\end{equation}
    
Here we further assume that the objective function is a
computationally-expensive-to-evaluate black box. For a general review of robust
optimisation with expensive functions see \cite{chatterjee2019critical}. When
the objective function is expensive to evaluate, optimisers such as
evolutionary algorithms \cite{branke1998creating, paenke2006efficient} or
particle swarm optimisation \cite{dippel2010using} will not be viable due to
the large number of function evaluations they demand. Therefore it is essential
to apply methods that necessitate only small numbers of observations.  In spite
of this requirement, relatively few methods exist in the literature to address
this problem \cite{chatterjee2019critical}.

A related field is that of \textit{level set estimation}
\cite{gotovos2013active}, where the aim is to determine a set of points of
equal objective function value, which encompass a region where the performance is guaranteed to
be better than some given threshold (either as a specific value or percentage
of the unknown optimum). \citet{bogunovic2016truncated} extended the idea
of level set estimation to work in a unified fashion with Bayesian optimisation.

There are a few methods in the literature that use GPs to develop a
surrogate model of the expensive function \cite{jin2011surrogate, ong2006max,
lee2006global}, which reduces the computational cost of the optimisation in two
ways. Firstly, it enables the surrogate model (rather than the expensive
function) to be searched using, for example, an evolutionary algorithm or
simulated annealing. Secondly, the use of a surrogate has the clear benefit of
curtailing the computational burden of evaluating the robustness of solutions,
because the surrogate model can be interrogated instead of the true objective
function. Although these methods lighten the computational load, they do not
take advantage of the uncertainty in the surrogate model, which is available
when using a GP and could be used to help guide the search in subsequent
iterations. \citet{picheny2013benchmark} present a review of robust acquisition
functions for use with a GP, but these only account for noise in the
function's response.

A state-of-the-art Bayesian approach was presented by
\citet{urrehman2014efficient}. This method exploits a GP with a modified
formulation of the expected improvement, which aims to account for the robust
performance over a region of the design space. Whilst this technique is shown
to be useful for expensive robust optimisation, there are two drawbacks:
\textit{(a)} the uncertainty of the GP is largely disregarded when
calculating the modified expected improvement, as only the uncertainty at the
estimated worst performing location is considered; and \textit{(b)} this method
is demonstrated with a somewhat substantial number of initial observations (100
in 10 dimensions), which makes it rather unsuitable for \textit{very} expensive
functions. Since this method is considered to be state-of-the-art we have
elected to include our own implementation of it for comparison during
experimentation.

More recently the StableOpt algorithm has been presented
\cite{bogunovic2018adversarially}.  This is a confidence-bounds approach that exploits a
Gaussian process model to locate a region of good inputs. In essence, at each
time-step $n$, a candidate robust location $\bx'$ is found as
\begin{align}
  \label{eq:stableopt-robust-location}
  \bx' = \argmin_{\bx \in \domain} \max_{\bdelta \in \robustset}
  \text{lcb}(\bx+ \bdelta)
  \,,
\end{align}
where $\text{lcb}(\cdot)$ denotes the lower confidence bound acquisition
function:
\begin{align}
  \label{eq:lcb}
  \text{lcb}(\bx) = \mu(\bx) - \sigma(\bx)
\end{align}
where $\mu(\bx)$ and $\sigma(\bx)$ are the posterior predictive mean and
variance at $\bx$ (equations \eqref{eq:pred-mean} and \eqref{eq:pred-variance}).
Adding  $\sigma(\bx)$ to $\mu(\bx)$ instead of subtracting it obtains the
upper confidence bound function, $\text{ucb}(\bx)$.
After identifying a candidate robust location, an expensive
evaluation is made at the location $\bx_{new}$ such that
\begin{align}
  \label{eq:stableopt-xnew}
  \bx_{new} = \argmax_{\bdelta \in \robustset} \, \text{ucb}(\bx + \bdelta).
\end{align}
Performing the search in this way ensures that the search operates
pessimistically when identifying robust optima, yet optimistically when
determining the next location at which to evaluate the expensive function.

%%
%% Bayesian search for robust optima.
%%

\section{Bayesian Search for a Robust Optimum}
\label{sec:bayesian-search-for-a-robust-optimum}

The search for a robust optimum can be distilled to the minimisation of the
selected quality function $Q(\bx, \robustset)$ (Section~\ref{subsec:robust_optimisation}) for a
given robust set $\robustset$. The obstacle to straightforward optimisation is
that evaluating $Q$ for any candidate location is unachievable for continuous
domains, because the evaluation of $f(\bx + \bdelta)$ is required for all
$\bdelta \in \robustset$. In spite of this, one can envisage that a good
approximation to $Q$ could be made by aggregating $f(\bx + \bdelta)$ over many $\bdelta \in
\robustset$. Although such an approach is practicable for cheap-to-evaluate
objective functions, the need to optimise expensive functions renders this
approximation infeasible as well. 

Here we propose to search for the robust optimum by constructing a Gaussian
process model of the expensive function $f$, and then using it to estimate the chosen
quality function $Q$ to determine the robust quality at a particular location.
This allows approximation of  the expected improvement for candidate
locations by drawing realisations from the Gaussian process and then evaluating
the improvement for each realisation based on these estimated quality values.
Algorithm~\ref{alg:bayesian_robust_optimisation} shows the main steps in our
robust optimisation procedure.

%%%
%Due to the expensive nature of our objective function we are restricted to only
%a small number of observations of the function $f$, which means that we must be
%intelligent when deciding where to evaluate the function at each iteration. Our
%suggestion is to evaluate the function at the location with maximum expected
%improvement, which we will approximate by drawing samples from the Gaussian
%process model.
%
%In order to account for the uncertainty in the modelled $f$ we
%approximate the expected improvement in $Q$ by drawing realisations from the GP
%and calculating the improvement for each realisation over the current best
%robust location. Then, averaging these realisation-specific improvements for
%all of the drawn realisations yields an approximation for the expected
%improvement.
%%%

\begin{algorithm}[t]
  \caption{Bayesian Robust Optimisation}
  \label{alg:bayesian_robust_optimisation}
  \begin{flushleft}
    {\bfseries Inputs}\\
    \setlength{\tabcolsep}{0pt}
    \hspace*{\algorithmicindent}\begin{tabular}[t]{ll}
      $X_N$:\quad\mbox{} &Initial $N$ observation locations\\
      $Y_N$: & Expensive evaluations  $f(X_N)$\\
      $w(\cdot)$: & Sampling function
                    (Section~\ref{subsec:sampling-location})\\
      $T$: &  Template of locations covering a robust set\\
      $\hat{Q}(F)$: & Function approximating quality of a set of function
                      evaluations $F$\\
      $M$: & Number of realisations to draw from GP\\
    \end{tabular}
    \bigskip
    
   {\bfseries Procedure}
   \end{flushleft}
  \begin{algorithmic}[1]
    \For{$n \gets N, N+1, \ldots$}
      \State{%
        Fit GP to $\{(\bx, y)\}_{X_n, Y_n}$
        \label{alg:gpmodel}
        \Comment{
          Maximise marginal likelihood \eqref{eq:log-marginal-likelihood}
        }
      }
      \State{%
        $\bestx_n \gets \mbox{best-so-far}(X_n)$}
        \label{alg:best_so_far}
        \Comment{Best-so-far robust location containing an evaluated
          $\bx_n$ \eqref{eq:estimated-best}}
      \State{%
        $\bx' \gets
        \argmax_{\bx \in \mathcal{X}} EI_\epsilon(\bx, \bestx_n, T)$ 
        \label{alg:best_location}
      }
      \State{%
        $\bx_{n+1} \gets w(\bx')$
        \label{alg:sample}
      }
      \State{%
        $X_{n+1} \gets X_n \cup \{\bx_{n+1}\}$
        \label{alg:update_observations}
        \Comment{Update set of observations}
      }
      \State{%
        $Y_{n+1} \gets Y_n \cup \{f(\bx_{n+1})\}$
        \label{alg:expensive_evaluation}
        \Comment{Expensively evaluate $\bx_{n+1}$}
      }
    \EndFor
    \State{%
      \Return $\mbox{best-so-far}(X_{n+1})$
    }

    \bigskip

    \Function{$EI_\epsilon$}{$\bx, \bestx, T$}
    \label{alg:ei_start}
    \Comment{Evaluate expected improvement at $\bx$}
    \State{%
      $X^\star_\delta \gets \{ \bestx + \bdelta_i \given \bdelta_i \in T\}$
      \Comment{Samples in $\robustset$ referred to $\bestx$}
    }
    \For{$m \gets 1, 2, \ldots, M$} \label{alg:draw_start}
      \State{%
        $X_\delta \gets \{ \bx + \bdelta_i \given \bdelta_i \in T\}$
        \Comment{Samples in $\robustset$ referred to $\bx$}
      }
      \State{%
        $\bK_{ij} \gets k(\bx_i, \bx_j)$ for all $\bx_i, \bx_j \in
        X_\delta \cup X^\star_\delta$
        \Comment{%
          Posterior covariance for locations in $X_\delta$ and $X_\delta^\star$
        }
      }
      \State{%
        $F \sim \normal(\bmu, \bK)$
        \label{alg:joint_sample}
        \Comment{
          Joint sample of a realisation across candidate $X_\delta$ and
          best-so-far $X_\delta^\star$
        }
      }
      
      \State{%
        $I_m \gets \max
        \left(0, 
          \hat{Q}(\{F_i \given F_i \in F \wedge \bx_i \in
          X_\delta^\star\}) - \hat{Q}(\{F_i \given F_i \in F \wedge \bx_i \in
          X_\delta\})
        \right)$
        \label{alg:improvement}
        \Comment{Improvement}
        \Statex\Comment{for this realisation}
      }
    \EndFor \label{alg:draw_end}
    \State{%
      \Return{$\frac{1}{M}\sum_{m=1}^M I_m$}
      \label{alg:expected_improvement}
    }
    \EndFunction
  \end{algorithmic}
\end{algorithm}

The process is initialised with a small number $N$ of evaluations of $f(\bx)$,
usually chosen using a low discrepancy sampling scheme
\cite{matousek1998l2discrepancy}, such as a Sobol' sequence
\cite{sobol1967distribution} or Latin hypercube sampling
\cite{morris1995exploratory,mckay1979comparison}. These allow a Gaussian
process to be constructed (line~\ref{alg:gpmodel}). 

In Algorithm~\ref{alg:bayesian_robust_optimisation} we make use of a
\textit{template} $T$ which is a discretised representation of the robust set
$\robustset$. The exact structure of $T$ is left to the practitioner, but there
are several considerations for constructing a template that would be useful in
practice. Firstly the set of points constituting the template could be
distributed uniformly within the bounds of the robust set, alternatively it
would be possible to arrange the set of points to be more densely distributed
near the centre of the robust set, thus giving greater weight to perturbations
closer to the centre of the robust set; cf equation \eqref{eq:quality-average}.
Finally, for worst case performance guarantees (equation \eqref{eq:quality-worst}) one might consider placing a
larger proportion of points on the boundary of the robust set, because this is
where one might assume the extremes of the function response within the robust
set would be.

Next, an appropriate quality function
(Section~\ref{subsec:robust_optimisation}) must be selected to estimate the
robust quality of a location. Again, the choice of this function is left to the
practitioner, but here we provide two such examples of modified quality
functions which operate on sets of response values for each location within the
robust set template. Denoting the set of responses estimated from the GP
surrogate by $F$, worst-case quality function is estimated as 
\begin{align}
  \hat{Q}_{max}(F) = \max_{F_i \in F} F_i
  \label{eq:estimated_qmax}
\end{align}
and the average quality function by
\begin{align}
  \hat{Q}_{avg}(F) = \frac{1}{|F|} \sum_{F_i \in F} F_i \, .
  \label{eq:estimated_qavg}
\end{align}

Over lines \ref{alg:draw_start} to \ref{alg:draw_end} of Algorithm \ref{alg:bayesian_robust_optimisation},  a realisation $f_m$ is
drawn from the fitted Gaussian process model for the set of $\bx$ that
constitute the candidate and best-so-far templates for the robust region
($X_\delta$ and $X^\star_\delta$ respectively). Note that a realisation
evaluated at a set of locations $\{\bx_k\}$ is a draw from a multivariate
Gaussian $\normal(\bzero, \bK)$ where $K_{kl} = k(\bx_k, \bx_l)$. Then the
improvement for each realisation can be calculated as per line
\ref{alg:improvement} and the expected improvement is the mean of the $M$
realisation-specific improvements (line~\ref{alg:expected_improvement}).

Ordinarily, in non-robust Bayesian optimisation, the best returned location is
one of the observations $\bx_n \in X_N$, which provides some guarantee about
the quality of the returned location. In order to provide a similar guarantee
we suggest that it is reasonable to enforce that the returned best robust
location should be within the vicinity of at least one expensively evaluated
location, that is for $\bestx$ to be considered a valid solution there must
exist $ \bx_n
\in X_N$ such that $d(\bestx - \bx_n) \leq \epsilon$, where $X_N$ is the set of
$N$ evaluated locations, and $d(\cdot)$ and $\epsilon$ are the distance
function and robustness parameter defining the robust set
\eqref{eq:robust_set_definition}.
For convenience we write
\begin{align}
  \label{eq:neighbourhood}
  \robustset(\bx) = \left\{\bx+\bdelta \given \bdelta \in \robustset\right\} 
\end{align}
for the robust set located at $\bx$.
Then the requirement that the robust optimum should be near an evaluated location
means that the set of possible locations for the best robust optimum found so
far is $\nbhd(X_N) = \cup_{\bx_n \in X_N}\robustset(\bx_n)$, which is illustrated in
Figure~\ref{fig:neighbourhood}. With expensively evaluated locations $X_N$, the
best robust location is identified as:
\begin{align}
  \label{eq:estimated-best}
  \bestx = \argmin_{\bx \in \nbhd(X_N)}
  \hat{Q}(\{\mu(\bx + \bdelta) \given \bdelta \in T\}) \, ,
\end{align}
and can be found using, for example, an evolutionary optimiser to search
over the $X_N$ (Algorithm \ref{alg:bayesian_robust_optimisation}
line~\ref{alg:best_so_far}). We discuss the computation of $\bestx$ in more
detail in Section \ref{subsec:best-sweetspot}.

As we show below, although we demand that a robust location is within the
vicinity of an evaluated location, it can be advantageous to evaluate $f$ at a
location other than that with maximum expected improvement. In
Section~\ref{subsec:sampling-location} we explore a number of criteria for
choosing the location to evaluate. In
Algorithm~\ref{alg:bayesian_robust_optimisation} $f$ is expensively evaluated
at the location provided by the function $w(\cdot)$ (lines \ref{alg:sample}
and \ref{alg:expensive_evaluation}). This sequence is then repeated until
convergence is achieved or computational resources are exhausted. 

We now illustrate the full procedure with a toy example.

\subsection{Toy Example}
\label{subsec:toy-example}
We illustrate the procedure using the toy one-dimensional function
\begin{equation}
  \label{eq:toy-function}
  f(x) = \sin(3\pi x^3) - \sin(8\pi x^3) \,,
\end{equation}
for $x \in \domain = [0, 1]$. For simplicity we restrict the robust set to
be an interval, where the robustness parameter $\epsilon=0.1$ defines the span
of the interval, and the shape function is given by $d(\delta)=\|\delta\|$,
where $x + \delta \in \domain$.

\begin{figure}[t!]
  \centering
  \subfloat[]{
    \includegraphics[trim={0mm, 3mm, 0mm, 0mm}, clip]
    {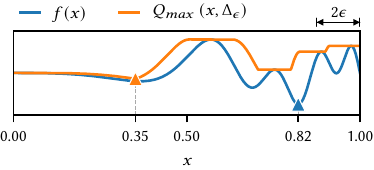}
    \label{fig:toy-and-induced-robust-landscape}
  }
  \subfloat[]{
    \includegraphics[trim={0mm, 3mm, 0mm, 0mm}, clip]
    {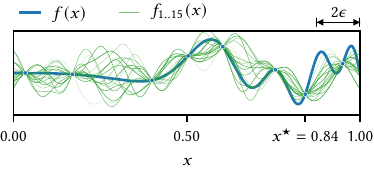}
    \label{fig:toy-gp-realisations}
  }

  \subfloat[]{
    \includegraphics[trim={0mm, 0mm, 0mm, 0mm}, clip]
    {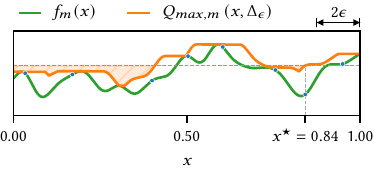}
    \label{fig:toy-realisation-improvement}
  }
  \subfloat[]{
    \includegraphics[trim={0mm, 0mm, 0mm, 0mm}, clip]
    {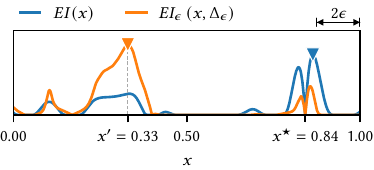}
    \label{toy-expected-improvement-comparison}
  }
  \caption{
    \textit{(a)} The toy function defined in \eqref{eq:toy-function} and the
    induced robust landscape using the worst-case quality function
    \eqref{eq:quality-worst}. The triangles (of respective colour) indicate the
    minimum of the toy and robust landscapes, and the bar in the top right-hand
    corner visualises the width of the uncertainty region.
    \textit{(b)} An example 15 realisations drawn from a Gaussian process,
    which has been fitted to 8 initial observations (dots) of the toy function
    $f$. The best---based on this Gaussian process model---robust location
    $\bestx$ is shown along the bottom.
    \textit{(c)} One of the realisations drawn in panel \textit{(d)} and the
    response of its corresponding quality function. The horizontal dashed line
    shows the quality of~$x_m^\star$ for \textit{this} realisation, and the
    shaded region indicates where there is improvement over the best-so-far
    location.
    \textit{(d)} Monte~Carlo approximation (using 100 realisations) of the
    robust expected improvement, and the usual single-point expected
    improvement \eqref{eq:ei_analytical}. Triangles (of respective colour)
    indicate where the expected improvement is greatest.
  }
  \label{fig:expected-improvement}
\end{figure}

Figure~\ref{fig:toy-and-induced-robust-landscape} shows the toy target function
$f$ and the induced robust landscape for the worst-case quality $Q_{max}$ as
given by \eqref{eq:quality-worst}. This toy function illustrates how the
optimal single point location, namely the minimum of $f(\bx)$, can exist in a
distinct location from the optimal robust region.

The first step is to fit a Gaussian process to an initial set of observations
of the expensive function $f$. In this instance we have used an initial set of
$N = 8$ observations; example draws from the resulting Gaussian process can be
seen in Figure~\ref{fig:toy-gp-realisations}.

Each realisation of the Gaussian process can be thought of as a potential
$f$ whose quality can be evaluated using the chosen quality function. Figure
\ref{fig:toy-realisation-improvement} shows the result of applying the quality
function $Q_{max}$ to a drawn realisation $f_m$ together with the resulting
improvement $I_m(x, \robustset)$ from that realisation.
For simplicity in this example, we have constrained the best uncertainty set to
be centred on an observation $x_n \in X_8$; the best uncertainty set so far
(for the depicted realisation) is centred at $x_m^\star = 0.842$, including the
observation at $x = 0.842$. The robust quality for each realisation is
calculated as the difference between the robust quality of the best-so-far
location and the candidate location; as shown in
Algorithm~\ref{alg:bayesian_robust_optimisation} on line~\ref{alg:improvement}
and the robust expected improvement is approximated as an average over all of
the realisations, using the procedure given between lines~\ref{alg:ei_start}
and \ref{alg:expected_improvement} of
Algorithm~\ref{alg:bayesian_robust_optimisation}. This is the acquisition
function used for determining where to sample $f$ next. Figure
\ref{fig:expected-improvement} (bottom) compares the robust expected
improvement with the (usual) single-point expected improvement
\eqref{eq:ei_analytical}, which clearly demonstrates that the robust expected
improvement gives greater weight to searching the more robust regions of design
space.
\begin{figure}[t]
  \centering
  \includegraphics[trim={0mm, 4mm, 0,mm 0mm}, clip]{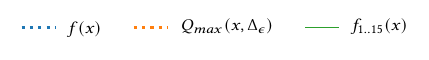}
  \subfloat{
    \includegraphics[width=0.5\textwidth,trim={0mm, 0mm, 0mm, 2mm}, clip]
    {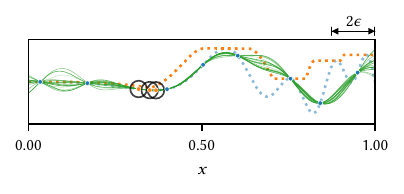}
  }
  \subfloat{
    \includegraphics[width=0.5\textwidth,trim={0mm, 0mm, 0mm, 0mm}, clip]
    {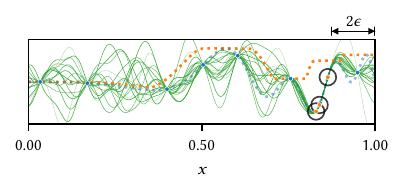}
  }
  \caption{
    Comparison of where the three observations following those in
    Fig.~\ref{fig:expected-improvement} are located when using the robust
    expected improvement (left) and the usual single-point expected improvement
    (right). The same eight initial observations were used for both schemes.
    The three new observations are indicated with circles. An example 15
    realisations, which were drawn after the new observations were made, have
    been depicted in both panels.
  }
  \label{fig:comparison-after-3-iterations}
\end{figure}
Figure~\ref{fig:comparison-after-3-iterations} shows the result of continuing
the optimisation procedure for three additional iterations; the objective
function is evaluated at the location of maximum expected improvement, $\bx'$.
The robust optimiser quickly locates the region of the robust optimum, whereas
the single-point optimiser searches the region of the (fragile) global minimum.

\subsection{Sampling Location}
\label{subsec:sampling-location}

Non-robust acquisition functions, such as the expected improvement described in
Section \ref{subsec:bayesian-optimisation}, determine a \textit{point} of
maximum acquisition. In contrast, their robust counterparts yield a region,
which presents an additional decision in the optimisation process: where within
this region should the next observation of the target function be made?

\begin{figure}[t!]
  \centering
  \includegraphics{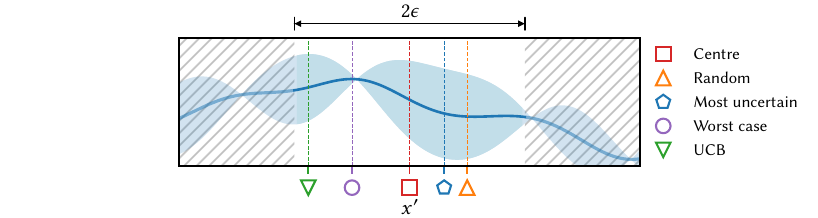}
  \caption{
    Illustration of each sampling function described in Section
    \ref{subsec:sampling-location}. The solid line and accompanying shaded
    region depict a fitted Gaussian process model and 95\% confidence interval.
    The hatched region indicates $\domain \setminus \robustset(\bx')$; the
    sampling function may only choose $\xnext \in \robustset(\bx')$. Note that
    the location depicted for the ``random'' sample is illustrative for one
    possibility of such a function.
  }
  \label{fig:sampling-location}
\end{figure}

We constrain the location of the new observation $\xnext$ to be within the
robust set of maximum acquisition $\robustset(\bx')$. Further, we propose
the use of a \textit{sampling function} $w(\bx')$, which determines where
within $\robustset(\bx')$ to locate the next expensive evaluation: $\xnext = w(\bx')$.
As we show empirically in section \ref{sec:results} on a range of test functions,
the choice of $w$ has a significant impact on the algorithm's ability to
converge to the best uncertainty region. Here we present five suggestions for
the sampling function $w$; Figure \ref{fig:sampling-location} illustrates each
of them.

\begin{description}[wide,leftmargin=30pt,labelsep=15pt]
  \item[Centred observation.] Usually uncertainty sets $\S(\bx)$ are
    symmetrical about $\bx$ and  an obvious choice is to observe the
    objective function at the location of maximum expected improvement---the
    centre of the uncertainty set:
    \begin{align}
      \label{eq:sampling-at-centre}
      w(\bx') = \bx'.
    \end{align}

  \item[Most uncertain observation.] A maximally explorative approach
    to improving the estimate of the quality of the predicted best
    uncertainty set
    is to observe the expensive function at the location of maximum uncertainty
    within $\robustset(\bx')$:
    \begin{equation}
      \label{eq:sampling-at-uncertain}
      w_{\sigma^2}(\bx') = \argmax_{\bx \in \robustset(\bx')} \sigma^2(\bx) \,,
    \end{equation}
    where $\sigma^2(\bx)$ is the predicted variance of the Gaussian process
    at $\bx$, see \eqref{eq:pred-variance}.
        
  \item[Worst-case prediction.] An alternative to improve the estimate
    of the uncertainty set's quality is to query at the location of the worst-case
    predicted value:
    \begin{equation}
      \label{eq:sampling-at-worstcase}
      w_{\mu}(\bx') = \argmax_{\bx \in \robustset(\bx')} \mu(\bx) \,,
    \end{equation}
    where $\mu(\bx)$ is the predicted mean of the Gaussian process at
    $\bx$.  This strategy has the benefit of confirming or revising the
    predicted worst case performance with an actual evaluation. 

  \item[Uniformly at random.] Finally, draw  $\xnext$ uniformly at
    random within $\robustset(\bx')$:
    \begin{align}
      w_{\mathcal{U}}&(\bx') = \text{random}(\robustset(\bx')) \,.
      \label{eq:sampling-at-random}
    \end{align}
    This approach may also be expected to promote exploration, but not in
    such a directed way as the ``most uncertain observation'' scheme.

  \item[UCB.] Taking inspiration from StableOpt, we locate  $\xnext$
    such that it maximises the upper confidence bound within the proposed
    robust region:
    \begin{align}
      w_{UCB}&(\bx') = \argmax_{\bx \in \robustset(\bx)} \mu(\bx) + \beta\sigma(\bx)\,
      \label{eq:sampling-at-ucb}
    \end{align}
    where, as with StableOpt, we take $\beta=2$.
  \end{description}

\subsection{Best-So-Far Robust Location}
\label{subsec:best-sweetspot}

\begin{figure}[t]
  \centering
  \includegraphics{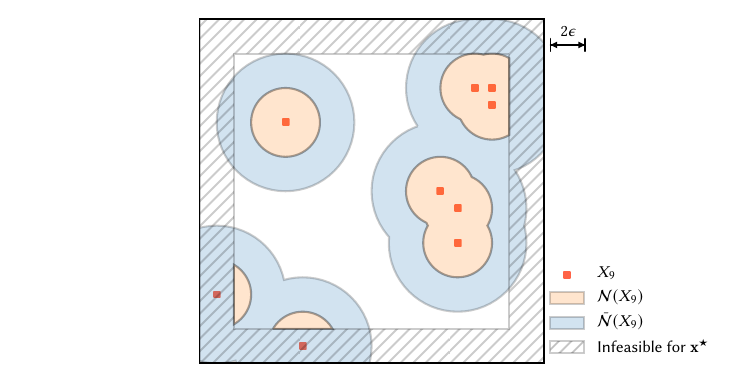}
  \caption{Illustration of the neighbourhood $\nbhd(X_N)$
    \eqref{eq:neighbourhood} and the extended neighbourhood
    $\bar\nbhd(X_N)$ \eqref{eq:extended-neighbourhood} for $N = 9$ example
    observations (red squares) in two dimensions. This set of observations
     demonstrates two consequences of the two
    neighbourhoods. Firstly, neither neighbourhood relies on the response
    value of the expensive function $f$. Secondly, because we demand that
    the entire robust region $\robustset$ exists within the domain
    $\domain$, it can be seen that there is a margin around the boundary of
    $\domain $ that prevents the robust
    region from existing too close to the edge of the domain. Note that new
    observations may be made within this margin.
    }
  \label{fig:neighbourhood}
\end{figure}

In standard Bayesian optimisation the best-so-far location $\bestx$ and its
function value $f(\bestx)$ are simply available because $f$ has been evaluated
at all $\bx \in X_N$, so deciding on the current $\bestx$ is merely a matter of
comparing the evaluated locations and then selecting the best one.  However, in
this robust scheme the improvement for a particular realisation $f_m$ requires
a procedure to search for the quality of the best robust region: $\max_{\bestx
\in \nbhd(X_N)}Q(\robustset(\bestx))$ so that candidate robust centres can be compared
with it. Since the evaluation of $Q(\robustset(\bx))$ requires evaluating the modelled
$f$ at a set of locations covering $\robustset(\bx)$, this optimisation in turn
requires evaluating the modelled $f$ over all locations that might be covered
by the robust region, that is over the extended neighbourhood of $X_N$:
\begin{align}
  \label{eq:extended-neighbourhood}
  \bar{\nbhd}(X_N) = \bigcup_{\bx \in \nbhd(X_N)} \robustset(\bx) \,.
\end{align}
The extended neighbourhood is illustrated in Figure \ref{fig:neighbourhood}.

To avoid this potentially expensive optimisation for every draw of a
realisation of $f$, we instead identify the best robust region as:
\begin{equation}
    \label{eq:best-sweetspot-centre}
    \bestx =
    \argmin_{\bx \in \nbhd(X_N)} \hat{Q} \left( \robustset(\bx) \right) \,,
\end{equation}
where $\hat{Q}$ is evaluated from the mean of the modelled $f$:
\begin{align}
  \label{eq:Qhat}
  \hat{Q}(\robustset(\bx)) = \max_{\bx' \in \robustset(\bx)} \mu(\bx') \,.
\end{align}
As shown in Algorithm~\ref{alg:bayesian_robust_optimisation}, $\bestx$ is
determined once each new observation is acquired (line
\ref{alg:best_location}).

By evaluating $f_m$ at a number of locations $\bx_k$ in a candidate robust set,
the improvement for a particular realisation is then evaluated as
\begin{align}
  \label{eq:estimated-improvement}
  I_m(\robustset(\bx)) =
    \max\left(0, Q_m^\star - \max_{\bx_k \in \robustset(\bx)} f_m(\bx_k)\right)
\end{align}
with the best quality found so far estimated as:
\begin{equation}
    \label{eq:estimated-Qjstar}
    Q_m^\star = \max_{\bx \in \robustset(\bestx)} f_m(\bx) \,.
\end{equation}

\subsection{Convergence}
\label{subsec:convergence}

Essentially we aim to perform Bayesian optimisation on the induced robust
landscape $Q(\bx)$ \eqref{eq:quality-worst}. As a result all of the usual
theoretical guarantees for the convergence of Bayesian optimisation already
presented in the literature for Bayesian optimisation apply. In particular
\citet{bull2011convergence} shows that Bayesian optimisers using the
expected improvement acquisation function as here converge under certain
conditions; see also \citep{vazquez2010convergence}.

However, it should be recognised that there are two points where this
algorithm makes approximations that affect guarantees of convergence, which
we discuss briefly here.

Firstly,  the expected improvement is approximated by averaging over sample
realisations drawn from the  Gaussian process modelling $f(\bx)$.   In the
large sample limit this approximation converges to the desired value, like
$O(1/M)$ in the number of realisations $M$.  However, in practice the
acquisition function is approximated with relatively few samples (up to
$M=1000$ in the experiments reported here.)

Secondly, we approximate the quality of a robust set $Q(\robustset(\bx))$ by
evaluating the Gaussian process at a set of locations over the template
covering the robust set (here we use 60 samples in 2 dimensions, 250 samples in
5 dimensions, and 400 samples in 10 dimensions). Clearly, this approximation
may under-estimate the worst case quality and it relies on the fidelity of the
Gaussian process approximation. Sufficiently dense sampling of
$\robustset(\bx)$ can achieve a good approximation and an alternative is to use
a search procedure such as \textsc{direct} \cite{jones1993lipschitzian} which
can provide an upper bound to the worst-case performance. In practice, however,
we have found this to be exorbitantly expensive.

\section{Evaluation}
\label{sec:results}

\begin{figure*}[t!]
  \centering
  \captionsetup[subfigure]{labelformat=empty}
  \subfloat[$f_1\,,$ Bumped Bowl]{
    \includegraphics[trim={2mm, 0mm, 0mm, 2mm}, clip]
    {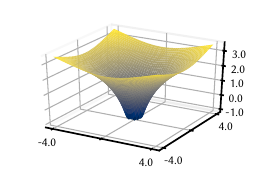}
    \label{fig:bumpedbowl3d}
  }
  \subfloat[$f_2\,,$ Levy 03]{
    \includegraphics[trim={2mm, 0mm, 0mm, 2mm}, clip]
    {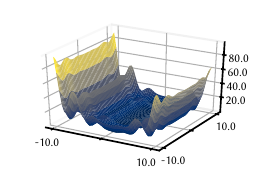}
    \label{fig:levy033d}
  }
  \subfloat[$f_3\,,$ Styblinski-Tang]{
    \includegraphics[trim={2mm, 0mm, 0mm, 2mm}, clip]
    {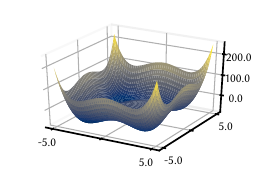}
    \label{fig:styblinskitang}
  }

  \subfloat[$f_4\,,$ Robust Problem 04]{
    \includegraphics[trim={2mm, 0mm, 0mm, 2mm}, clip]
    {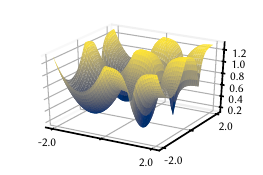}
    \label{fig:robustproblem43d}
  }
  \subfloat[$f_5\,,$ Stepped Sphere]{
    \includegraphics[trim={2mm, 0mm, 0mm, 2mm}, clip]
    {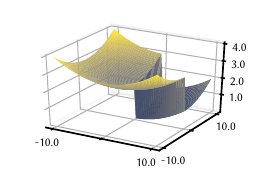}
    \label{fig:steppedsphere3d}
  }
  \subfloat[$f_6\,,$ Quintic]{
    \includegraphics[trim={2mm, 0mm, 0mm, 2mm}, clip]
    {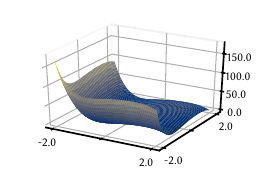}
    \label{fig:quintic3d}
  }
  \caption{
    Two-dimensional visualisations of the six benchmark functions. Exact
    formulations can be found in Table \ref{tab:benchmarks}.
  }
  \label{fig:benchmarks}
\end{figure*}

We present results of the performance of our method in comparison to the
state-of-the-art methods StableOpt \cite{bogunovic2018adversarially} and the
method described by \citet{urrehman2014efficient} with $D \in \{5, 10\}$ over
six common benchmark functions \cite{mirjalili2016obstacles,
laguna2005experimental, styblinski1990experiments}. We also include some
visualisations of results in two dimensions, which serve to demonstrate the
differences in search regimes between the compared methods. Figure
\ref{fig:benchmarks} presents two-dimensional visualisations of each function
for reference, which are defined in Table \ref{tab:benchmarks}.
          
\begin{table*}[!t]
  \centering
  \caption{
    Summary of the benchmark functions used for experimentation, the domains
    $\domain$ over which they were evaluated. A $\dagger$ signifies that the
    stated equation has been modified from the referenced source (see Section
    \ref{sec:results} for full explanation).
  }
  \label{tab:benchmarks}
  \begin{tabular}{ccc}
    \toprule
    Name & Equation & $\domain$ \\
    \midrule

    \makecell{Bumped Bowl$^\dagger$ \\ \cite{mirjalili2016obstacles}} &
    \( f_1(\bx) = \log \|\bx\|^2 + e^{-10 \|\bx\|^2} \) &
    $[-4, 4]^D$ \\
    \hline
    \makecell{Levy03 \\ \cite{laguna2005experimental}} &
    {$\!\begin{aligned}
      f_2&(\bx) = \sin^2(\pi x_1) + \sum_{d=1}^{D-1}
      (\omega_d - 1)^2 [1 + 10\sin^2(\pi \omega_{d+1})]\\
      & + (\omega_D - 1)^2 [1 + \sin^2(2\pi \omega_D)]\
      \text{; } \omega_d = 1 + {(x_d - 1)}/{4}
    \end{aligned}$} &
    $[-4, 4]^D$ \\
    \hline

    \makecell{Styblinski-Tang \\ \cite{styblinski1990experiments}} &
    \(
      f_3(\bx) = \frac{1}{2} \sum_{d=1}^{D} \left(x_d^4 - 16x_d^2 + 5x_d\right)
    \) &
    $[-5, 5]^D$ \\
    \hline

    \makecell{Robust Problem 4 \\ \cite{mirjalili2016obstacles}} &
    \makecell{
      \(f_4(\bx) = 1.3 - \frac{1}{d} \sum_{d=1}^{D} H(x_d)\,;\) \\
      \(
        H(x_d) = \begin{cases}
          -(x_d+1)^2 + 1,    & \text{if $x_d < 0$} \\
          2.6^{-8|x_d - 1|}, & \text{otherwise}
        \end{cases}
      \)
    } &
    $[-2, 2]^D$ \\
    \hline

    \makecell{Stepped Sphere$^\dagger$ \\ \cite{mirjalili2016obstacles}} &
    \makecell{
      \(
        f_5(\bx) = D-D\prod_{d=1}^{D}
        G(x_d)+\frac{1}{100}\sum_{d=1}^{D}x_d^2\,;
      \) \\
      \(
        G(x_d) = \begin{cases}
          1, & \text{if $x_d < $0} \\
          0, & \text{otherwise}
        \end{cases}
      \)
    } &
    $[-10, 10]^D$ \\
    \hline

    \makecell{Quintic \\ \cite{al-roomi2015}} &
    \(\displaystyle
      f_6(\bx) = \sum_{d=1}^{D} \left(x_{d}^5 - 3x_{d}^4 + 4 x_{d}^3 +
      2x_{d}^2 - 10 x_{d} - 4 \right)
    \) &
    $[-10, 10]^D$ \\
    \bottomrule
  \end{tabular}
\end{table*}

Each benchmark was selected to test a different aspect of robust optimisation.
Function $f_1$ (Bumped Bowl) presents a situation where the robust optimum is situated at a
local maximum, which tests the ability of an algorithm to overlook the
better-performing non-robust region. Our implementation of this benchmark has
been modified from \cite{mirjalili2016obstacles} to ensure that the robust
optimum exists exactly at the peak of the local maximum. Benchmarks Levy 03
and Mirjalili \& Lewis 04 ($f_2$and
$f_4$) are examples of functions with multiple local minima. In the case of
benchmarks $f_3$ (Styblinkski-Tang) and $f_6$ (Qunitic), the robust optimum resides just outside of the
global optimum, which tests robust procedures' resilience to non-robust
regions. Optimisers that exploit the parabolic sphere have difficulty in
finding the ``step'' in $f_5$ (Stepped Sphere) containing the optimum, which occupies a
vanishingly small proportion of the domain as the number of dimensions
increases. This function has been modified from \cite{mirjalili2016obstacles}
to ensure that the size of the step remains significant as $D$ increases: even
so, the proportion of $\domain$ containing the lower step is only $2^{-D}$.

\begin{figure*}[t!]
  \centering
  \includegraphics{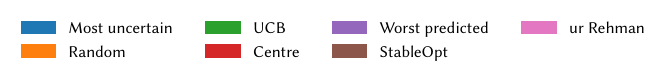}
  \includegraphics[width=0.96\linewidth]{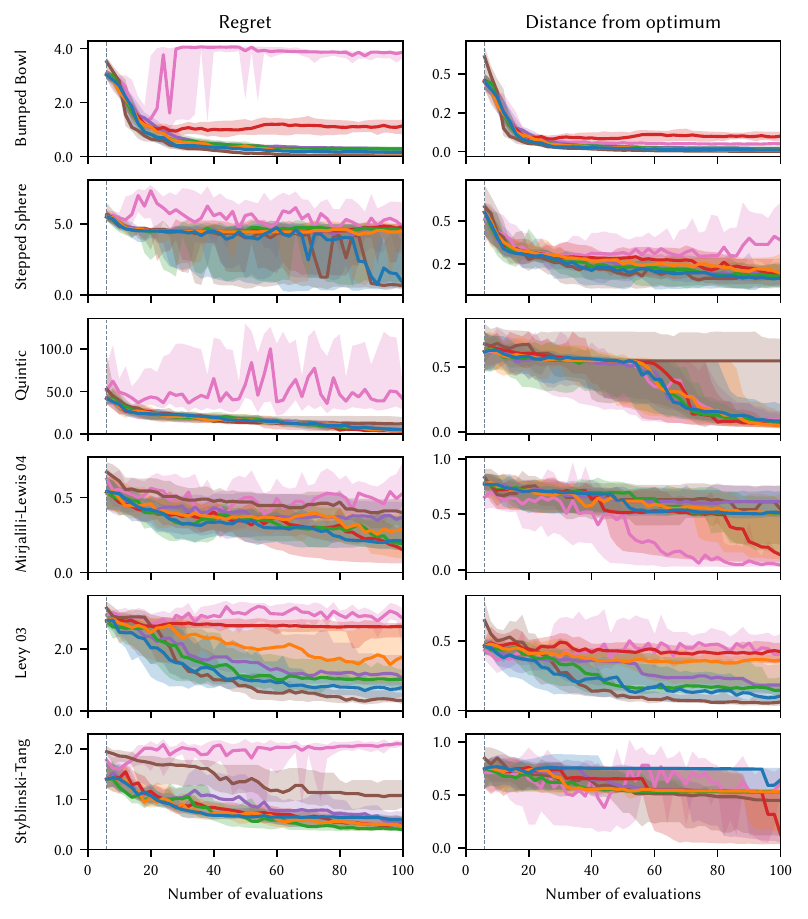}

  \caption{
    Results from 30 repeated trials on the six benchmarks in 5 dimensions
    (Section~\ref{sec:results} and Table~\ref{tab:benchmarks}) of each sampling
    scheme (given in Section \ref{subsec:sampling-location}), StableOpt
    \cite{bogunovic2018adversarially}, and the scheme described by
    \citet{urrehman2014efficient}. In the left-hand plots the solid lines
    show the median regret, and the shaded regions show the inter-quartile
    range. The right-hand plots show the distance between the proposed robust
    optimum and the true robust optimum. Vertical dashed lines indicate the
    initial Latin hypercube samples. The distances have been normalised to be
    within a unit cube domain.
  }
  \label{fig:convergence-plots-5d}
\end{figure*}

\begin{figure*}[t!]
  \centering
  \includegraphics{./figures/legend.pdf}
  \includegraphics[width=0.95\linewidth]{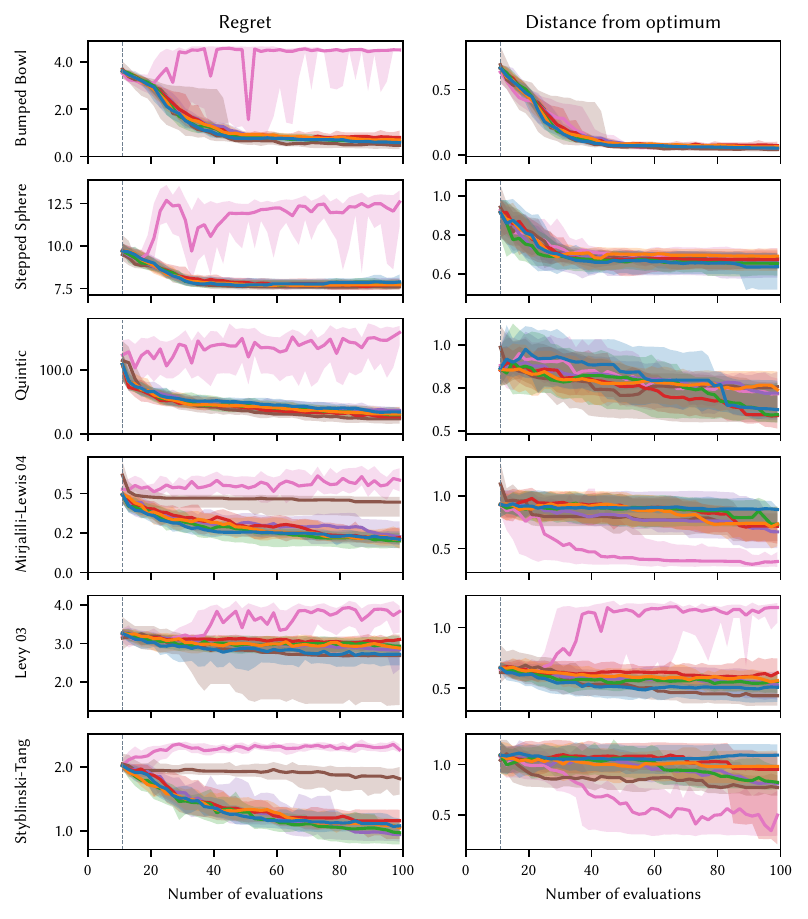}

  \caption{
    Results from 30 repeated trials on the six benchmarks in 10 dimensions
    (Section~\ref{sec:results} and Table~\ref{tab:benchmarks}) of each sampling
    scheme (given in Section \ref{subsec:sampling-location}), StableOpt
    \cite{bogunovic2018adversarially}, and the scheme described by 
    \citet{urrehman2014efficient}. In the left-hand plots the solid lines
    show the median regret, and the shaded regions show the inter-quartile
    range. The right-hand plots show the distance between the proposed robust
    optimum and the true robust optimum. Vertical dashed lines indicate the
    initial Latin hypercube samples.
  }
  \label{fig:convergence-plots-10d}
\end{figure*}

The robust quality measure used for evaluating candidate locations was the
worst-case quality $Q_{max}$ \eqref{eq:quality-worst}, the elected shape
function was $d(\bdelta)=\|\bdelta\|_2$, and the robustness parameter
$\epsilon= \frac{|u-l|}{8}$, where $u$ and $l$ are the upper and lower bounds
of the domain respectively (i.e. $\domain=[u, l]^D$).  See supplementary
material \ref{sec:5-dimensional} and \ref{sec:10-dimensional} for comparative
results using a ``square-shaped'' robust region $d(\bdelta)=\|\bdelta\|_1$ in 5
and 10 dimensions respectively.

As noted above the functions were modelled with a Gaussian process with a
Mat\'ern 5/2 kernel; kernel parameters were inferred by optimising the marginal
likelihood \eqref{eq:log-marginal-likelihood} using L-BFGS from 10 random
restarts. $M \in \{100, 500, 1000\}$ realisations were used in the evaluation
of the acquisition function, where $M$ increases if no improvement is found at
lower values. The location of the acquisition function maximum was found by
evaluating it for 1000 Latin hypercube samples and then optimising the most
promising 10 of these with L-BFGS-B \cite{byrd:lbfgs}; the optimisation budget
was 100 evaluations of the expensive function.

Python code to generate figures and reproduce all experiments is available
online\footnote{\url{https:github.com/url/completed/on/publication}}

We evaluated the five sampling schemes proposed in Section
\ref{subsec:sampling-location}: centred observation
\eqref{eq:sampling-at-centre}, most uncertain observation
\eqref{eq:sampling-at-uncertain}, maximised UCB \eqref{eq:sampling-at-ucb}, and
uniformly at random \eqref{eq:sampling-at-random}. 

To enable paired comparisons, each method was initialised using the same
$D+1$ Latin hypercube samples. The experiments were repeated 30 times for
statistical comparison. Figures~\ref{fig:convergence-plots-5d} and
\ref{fig:convergence-plots-10d} compare the convergence of each of the
methods tested. For the functions $f_2$, $f_3$, and $f_6$, where it is
difficult to analytically determine the robust optimum, we have completed
20 repeated trials of CMA-ES \cite{hansen2003reducing} in order to locate an
approximation for the global robust optimum.

%%
%% Analysis of results.
%%

\subsection{Analysis}
\label{subsec:analysis}

The convergence plots in Figure~\ref{fig:convergence-plots-5d} and
Figure~\ref{fig:convergence-plots-10d} show the regret, namely the difference between the state of
the optimiser and the value of the true robust minimum.  They demonstrate that
our robust optimisation procedure is generally capable of locating and
exploiting robust optima with a small number of observations of the underlying
expensive function.  However, we note that all methods perform significantly
less well in $D=10$ dimensions and none of the methods are able to locate a
good value for the robust optimum for the Levy03 or stepped sphere functions. 

\begin{figure}[t]
  \centering
  \subfloat[]{
    \includegraphics[width=\textwidth]{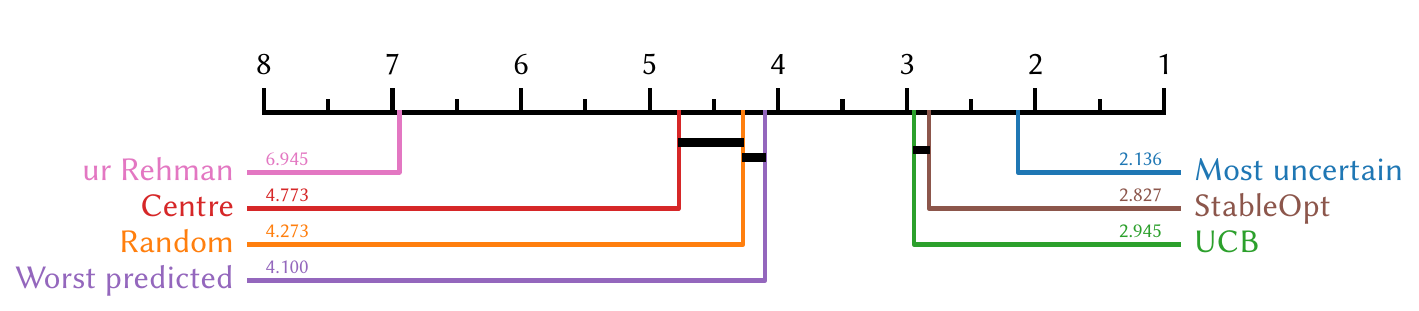}
    \label{fig:5d_cd_plot}
  }

  \subfloat[]{
    \includegraphics[width=\textwidth]{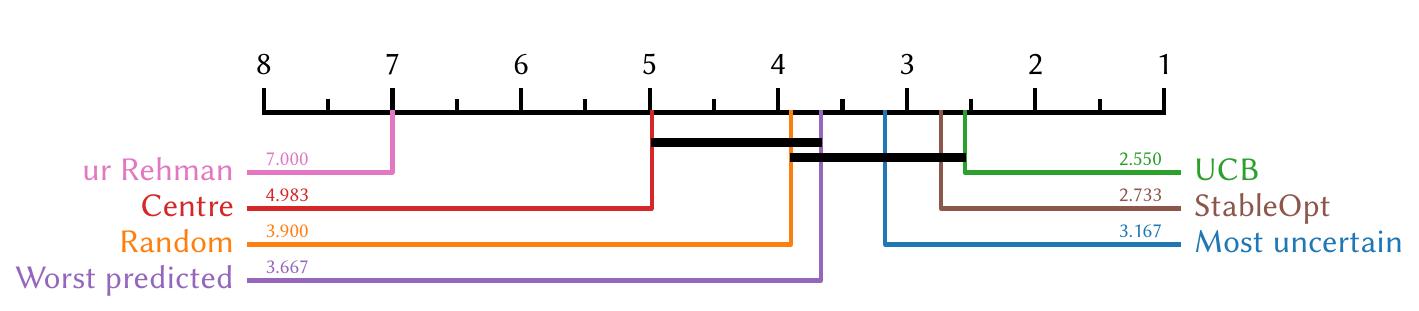}
    \label{fig:10d_cd_plot}
  }
  \caption{
    Critical difference plots \cite{demvsar2006statistical} over all functions
    for each of the five selection methods, StableOpt, and ur Rehman for the
    experimental results in  (a) five and (b) ten dimensions. The average ranks
    are shown on the horizontal axis (a lower rank is better), and the bold
    black horizontal lines connect algorithms that are statistically
    indistinguishable from one another using a Wilcoxon signed rank test
    \cite{wilcoxon1945individual}.
  }
  \label{fig:cd_plots}
\end{figure}

A summary of all of the tested algorithms over all of the functions is shown as
a critical difference plot \cite{demvsar2006statistical} in
Figure~\ref{fig:cd_plots}. Of the five competing sampling functions,
$w(\cdot)$, sampling at the most uncertain location within the robust region
consistently enables better convergence in $D = 5$ dimensions. In $D = 10 $
dimensions the performance of our algorithm with several sampling methods and
StableOpt are statistically indistinguishable using the Wilcoxon signed rank
test at $p = 0.05$.

The success of the most uncertain sampling strategy with $D=5$ is largely
because of the increased exploration of the best-so-far robust region,
which leads to more even coverage of observations in that key region. The
benefits of increased exploration are evident for the Stepped Sphere $f_5$
in $D=5$ dimensions where the most uncertain method is the only method of
ours that explores the bottom of the parabolic bowl containing the step
sufficiently to locate the optimum. In higher dimensions we conjecture that
the model of the function is not sufficiently good to allow identification
of the most uncertain point.

\begin{figure}[!t]
  \centering
  \subfloat[Most uncertain]{
    \includegraphics[trim={2mm, 0mm, 2mm, 0mm}, clip]{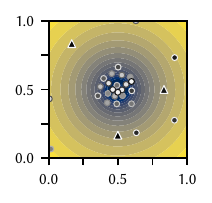}
    \label{fig:uncertain-sample-locations}
  }
  \subfloat[Random]{
    \includegraphics[trim={2mm, 0mm, 2mm, 0mm}, clip]{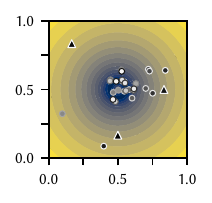}
    \label{fig:random-sample-locations}
  }
  \subfloat[UCB]{
    \includegraphics[trim={2mm, 0mm, 2mm, 0mm}, clip]{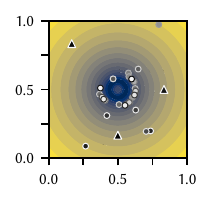}
    \label{fig:ucb-sample-locations}
  }
  \subfloat[Centre]{
    \includegraphics[trim={2mm, 0mm, 2mm, 0mm}, clip]{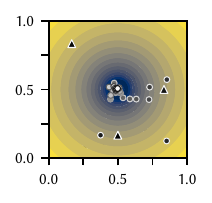}
    \label{fig:centre-sample-locations}
  }

  \subfloat[Worst predicted]{
    \includegraphics{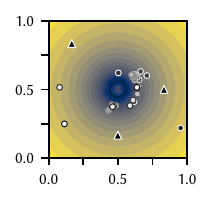}
    \label{fig:worst-sample-locations}
  }
  \subfloat[ur Rehman et al.]{
    \includegraphics{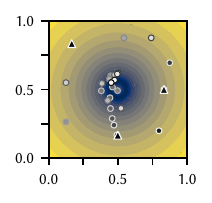}
    \label{fig:urrehman-sample-locations}
  }
  \subfloat[StableOpt]{
    \includegraphics{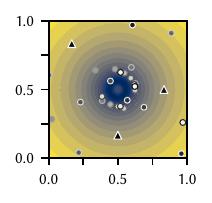}
    \label{fig:stableopt-sample-locations}
  }

  \caption{
    Comparison between where the expensive function ($f_1$) has been observed
    by the five sampling schemes proposed in Section
    \ref{subsec:sampling-location} \textit{(a)--(e)}, ur Rehman et al.'s method
    \textit{(f)}, and StableOpt \textit{(g)}. Triangles indicate the three
    initial samples, dots indicate subsequent observation locations. The colour
    of the markers lightens from black to white as the algorithm
    progresses.  Coordinates are normalised to $[0, 1]^2$ so that the
    robust optimum lies at $(\tfrac{1}{2}, \tfrac{1}{2})$ which is a local
    maximum of $f(\bx)$.
  }
  \label{fig:bumpedbowl-sample-locations}
\end{figure}

Figure~\ref{fig:bumpedbowl-sample-locations} shows the typical search pattern
for each of the five sampling methods, StableOpt,  and ur Rehman et al.'s
approach after 30 iterations on the two-dimensional Bumped Bowl function
\cite{mirjalili2016obstacles}. Each run was initialised from the same set of 3
Latin hypercube samples. It is clear from this example that both the ``most
uncertain'' sampling scheme has made a better estimate of the robust optimum.
In addition, the ``most uncertain'' sampling scheme has been exploratory over
the remainder of the domain, and has lead to the most even coverage of
observations at the robust optimum.

In all of the benchmark functions, our method has out-performed that of ur
Rehman et al.'s competing method, with respect to the value of regret at the
final iteration. It should be noted that ur Rehman et al.'s method was able to
get \textit{closer} to the optimum location on numerous occasions, but
struggled to accurately settle its robust region, which resulted in a much
worse value of regret (due to the fragility of the function landscape).

Generally, and as one might expect, the performance of all of the compared
schemes worsens as the number of dimensions increases. Benchmark problem $f_5$
is difficult for a GP to model due to the discontinuous downward step in one
corner of the domain. This problem is exacerbated in higher dimensions: whilst
in two dimensions the downward step covers $\frac{1}{4}$ of the domain, as
noted above, the step scales such that it occupies $\frac{1}{2^D}$ of the
domain. The result is that in ten dimensions the downward step exists in less
than one-thousandth of the domain. With the extremely limited number of
observations made during our experiments it is unsurprising that none of the
sampling schemes, StableOpt nor ur Rehman et al.'s method discovered the step.
In five dimensions only the ``most uncertain'' and StableOpt were able to find
the step, which indicates that the improved exploration offered by these
schemes presents significant advantages in locating complicated response
features.

In general ur Rehman et al.'s approach and our approach using the ``centre''
sampling scheme do not do well, and in fact perform similarly poorly in similar
circumstances. This appears to be a result of both of the approaches tending
towards making more exploitative observations, which means that they
repeatedly sample in approximately the same location and fail to construct
an accurate model of $f$.

Each of the presented methods make significant ground towards improving their
quality of robustness, as seen in Figures~\ref{fig:convergence-plots-5d} and
\ref{fig:convergence-plots-10d}. And each one generally exhibits a similar
convergences curve for the initial 20 iterations as the vicinity of the
robust optimum is approached.  However, during the remainder
of the run it is clear that ``most uncertain'' and ``random'' generally
outperform the other approaches.

\subsection{Real-world application}
\label{subsec:realworld}

\begin{figure}[t]
  \centering
  \subfloat[]{
    \includegraphics[width=0.3\linewidth]{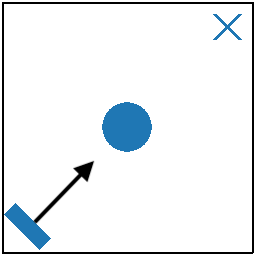}
    \label{fig:push4d_example}
  }
  \qquad
  \subfloat[]{
    \includegraphics[width=0.3\linewidth]{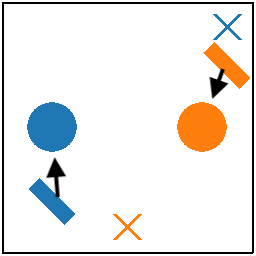}
    \label{fig:push8d_example}
  }
  \caption{
    Example of the initial setups for the robot pushing problems: Push4 (left)
    and Push8 (right). The rectangles are the robot hands, the circles are the
    objects to be pushed, and the crosses are the target locations for the
    objects. The black arrows indicate the direction of travel of the robot
    hands (always towards the object's initial position).
  }
  \label{fig:push_example}
\end{figure}

Here we illustrate our  robust optimisation method on a real-world problem. We
aim to optimise the input parameters for two active learning robot pushing
problems \cite{wang2018active}. In the first problem a single robot hand aims
to push an object towards an unknown target location; see Figure
\ref{fig:push4d_example}. Once the robot has pushed the object, per its input
parameters, the push is then evaluated as the distance between the target and
the final location of the pushed object. The input parameters to this problem
are the starting position of the robot hand, the orientation of the robot hand,
and the number of time steps for which the robot hand moves. The direction of
travel of the robot hand is constrained to be in the direction of the object's
initial location. We refer to this four-dimensional problem as \textit{Push4}.

In the second problem, as shown in Figure \ref{fig:push8d_example}, two robot
hands are tasked to push their respective object towards two unknown targets.
This problem poses an additional layer of complexity in that both of the robot
hands and objects may block each other's path if they intersect. The resulting
feedback from this problem is the sum of the distance between both of the
objects and their respective targets. Due to the possibility of the robots and
objects blocking each other, some problem instances do not have perfect
solutions. There are eight input parameters to optimise in this case: the
starting position of both of the robot hands, the orientation of both of the
robot hands, and the number of time steps for which each of the robot hands
move. The resulting eight-dimensional problem is referred to as \textit{Push8}.

As with \citep{wang2017max}, the object's initial location in Push4 is always
at the centre of the domain, but the target location is changed for each of the
optimisation runs. Each of the compared algorithms used the same target
locations, so that the runs are directly comparable. Doing this means that we
average over instances of the problem class, rather than repeatedly optimising
the same function from a different set of starting observations.  Likewise, in
Push8 the starting position of the objects were fixed, and their respective
targets were randomised for each optimisation run. In order to estimate the
optimum for each of the different initialisations of the problems we initially
sampled the feasible space with 1000 different robot parameters, and then
locally optimised the landscape using CMA-ES from the best 20 sampled points.

\begin{figure}[t]
  \centering
  \includegraphics{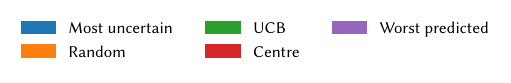}
  \includegraphics{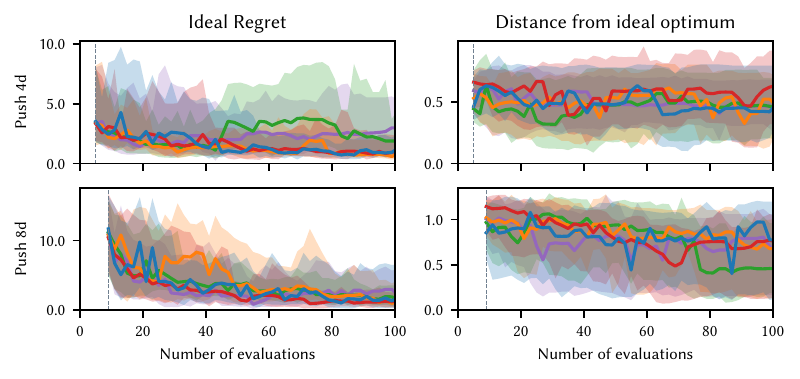}
  \caption{
    Results for a \textit{spherical} robust region over 30 repeated trials on
    the two real-world applications: Push4 and Push8. The left-hand plots
    depict the approximated regret, and the right-hand plots show the
    distance from an approximated optimum. The solid lines are the median over
    the 30 trials, and the shaded region is the inter-quartile range. The
    vertical dashed line indicated the initial Latin hypercube samples.
  }
  \label{fig:push_results}
\end{figure}

The results of the experiments, shown in Figure~\ref{fig:push_results}, show
that on the four-dimensional Push4 that the ``most uncertain'', ``random'', and
``centre'' selection methods achieve the best median value of regret. However,
in eight dimensions all of the selection methods are capable of improving the
function value, but there is little to discern between them. In fact all of the
selection methods are considered statistically indistinguishable. This is
perhaps unsurprising, because the landscape for the \textit{Push8} problem is
very complicated: there are many plateaux and discontinuities throughout the
landscape, which means that many of the problem instances are very difficult
for a Gaussian process to model. It should be noted that in spite of this
difficulty, all of our proposed selection methods make some headway to
improving the fitness value, and are actually able to find better results than
our initial estimates of the optima using random sampling and CMA-ES.

%%
%% Conclusion and future work.
%%

\section{Conclusion}
\label{sec:conclusion}

This paper has introduced a novel algorithm for the robust optimisation of
expensive-to-evaluate functions in the context of Bayesian optimisation.
Experiments on a range of commonly-used benchmark functions show that our
method is effective at locating robust optima, and able to outperform two
state-of-the-art methods from the literature. The algorithm depends upon
building a model of the expensive function and then evaluating the improvement
with respect to a chosen quality function over realisations drawn from the
model.  The expectation of these improvements is then used as an acquisition
function, which is maximised to inform the next location of the expensive
function to evaluate.

Standard (non-robust) Bayesian optimisation, StableOpt and ur Rehman et al.'s
method only require the GP to be evaluated once per iteration; we require $M$
evaluations of the GP to form an estimate of the robust improvement.
Consequently our methods take longer roughly $M$ times as long to decide on the
next location for evaluation of the expensive function. While this additional
burden is significant for benchmark functions like these which are trivial to
evaluate, the additional time required is insignificant in comparison with the
time required to evaluate real-world expensive functions.  Note also that
interrogating the GP can be made in parallel.

Subject to the demand that the expensive-to-evaluate function be evaluated in
the putative robust region, we have demonstrated that the choice of sampling
location can markedly affect the convergence rate and quality of the final
solution. Methods that promote exploration are generally more effective than
exploitative methods; sampling from the location about which the model is
most uncertain is effective, although we suspect that in higher dimensions that
quality of the surrogate model may not be good enough to properly identify the
most uncertain location.

% Future work.
Future work entails the simultaneous optimisation of the location and the shape
of the robust region, and improvements in uncertainty estimation in high
dimensions.

\begin{acks}
\label{sec:acknowledgement}
This work was supported by the Engineering and Physical Sciences Research
Council [grant number EP/M017915/1].
\end{acks}

% End matter.
\bibliographystyle{ACM-Reference-Format}
\bibliography{references} 

\beginsupplement

\section{Square robust region}
\subsection{5-dimensional results}
\label{sec:5-dimensional}
\begin{figure}[H]
  \includegraphics{./figures/legend.pdf}
  \includegraphics[width=0.85\textwidth]{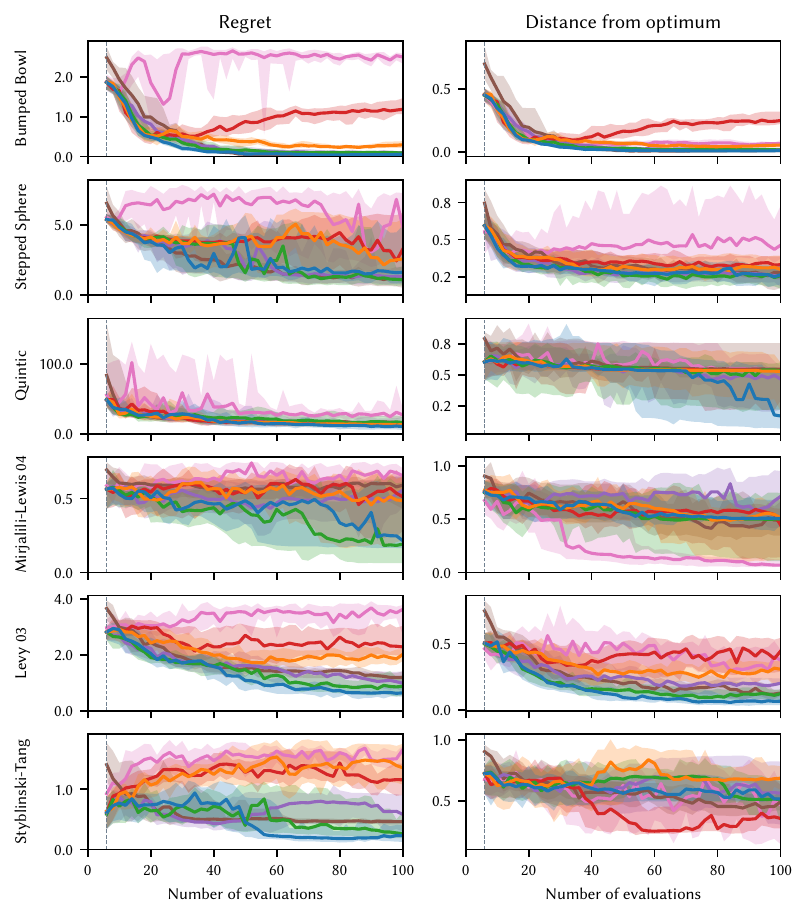}
  \caption{
    Results for a \textit{square-shaped} robust region over 30 repeated trials
    on the five benchmarks in 5 dimensions (Section~\ref{sec:results} and
    Table~\ref{tab:benchmarks}) of each sampling scheme (given in Section
    \ref{subsec:sampling-location}), StableOpt
    \cite{bogunovic2018adversarially}, and the scheme described by 
    \citet{urrehman2014efficient}. In the left-hand plots the solid lines
    show the median regret, and the shaded regions show the inter-quartile
    range. The right-hand plots show the distance between the proposed robust
    optimum and the true robust optimum. The vertical dashed line indicates the
    initial Latin hypercube samples.
  }
  \label{fig:5d_results_square}
\end{figure}
\begin{figure}[h!]
  \includegraphics[width=\textwidth]{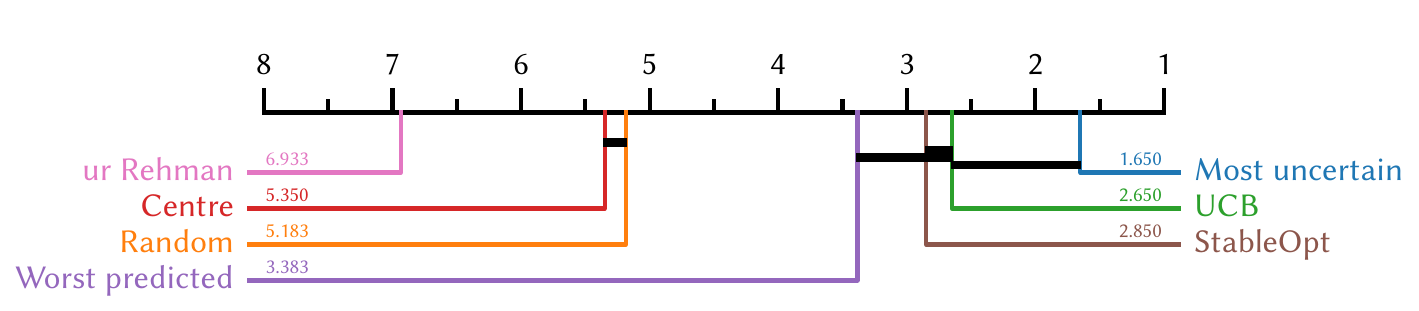}
  \caption{
    Critical difference plots \cite{demvsar2006statistical} over all functions
    for each of the five selection methods, StableOpt, and ur Rehman for the
    experimental results in five dimensions for a square-shaped robust region.
  }
  \label{fig:5d_cd_square}
\end{figure}

\subsection{10-dimensional results}
\label{sec:10-dimensional}
\begin{figure}[H]
  \includegraphics{./figures/legend.pdf}
  \includegraphics[width=0.9\textwidth]{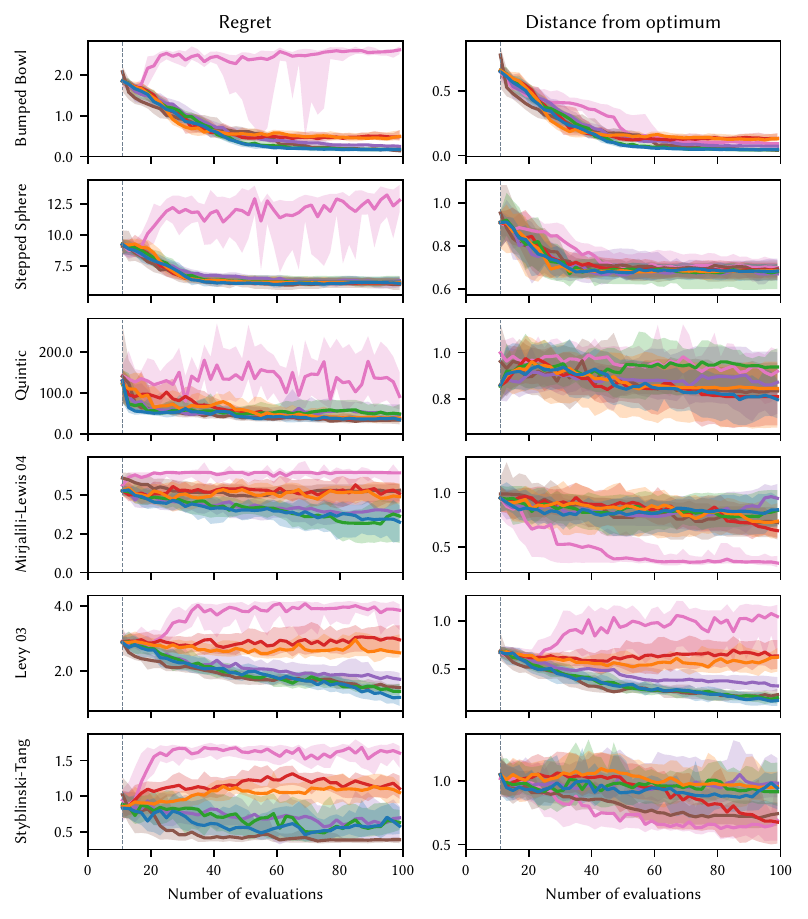}
  \caption{
    Results for a \textit{square-shaped} robust region over 30 repeated trials
    on the five benchmarks in 10 dimensions (Section~\ref{sec:results} and
    Table~\ref{tab:benchmarks}) of each sampling scheme (given in Section
    \ref{subsec:sampling-location}), StableOpt
    \cite{bogunovic2018adversarially}, and the scheme described by \citet{urrehman2014efficient}. In the left-hand plots the solid lines
    show the median regret, and the shaded regions show the inter-quartile
    range. The right-hand plots show the distance between the proposed robust
    optimum and the true robust optimum. The vertical dashed line indicates the
    initial Latin hypercube samples.
  }
  \label{fig:10d_results_square}
\end{figure}
\begin{figure}[h!]
  \includegraphics[width=\textwidth]{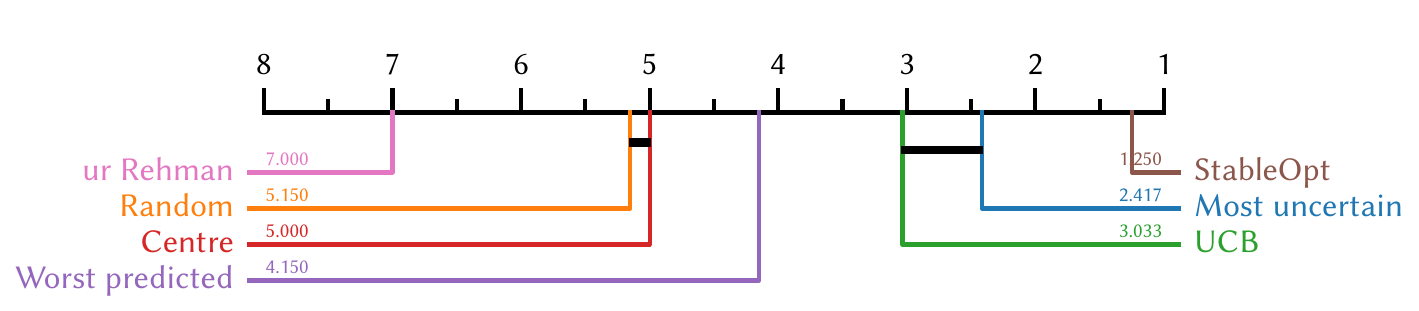}
  \caption{
    Critical difference plots \cite{demvsar2006statistical} over all functions
    for each of the five selection methods, StableOpt, and ur Rehman for the
    experimental results in ten dimensions for a square-shaped robust region.
  }
  \label{fig:10d_cd_square}
\end{figure}

\end{document}